\documentclass[11pt,a4paper]{article}
\usepackage[hyperref]{emnlp2020}
\usepackage{times}
\usepackage{latexsym}

\usepackage{microtype}
\usepackage{graphicx}

\aclfinalcopy %
\def\aclpaperid{1750} %

\title{Exploring the Role of Argument Structure in Online Debate Persuasion}

\author{Jialu Li \\
  Cornell University \\
  \texttt{jl3855@cornell.edu} \\\And
  Esin Durmus \\
  Cornell University \\
  \texttt{ed459@cornell.edu} \\\And
  Claire Cardie \\
  Cornell University \\
  \texttt{cardie@cs.cornell.edu} \\}

\date{}

\begin{document}
\maketitle
\begin{abstract}
Online debate forums provide users a platform to express their opinions on controversial topics while being exposed to opinions from diverse set of viewpoints. Existing work in Natural Language Processing (NLP) has shown that linguistic features extracted from the debate text and features encoding the characteristics of the audience are both critical in persuasion studies. In this paper, we aim to further investigate the role of discourse structure of the arguments from online debates in their persuasiveness. In particular, we use the factor graph model to obtain features for the argument structure of debates from an online debating platform and incorporate these features to an LSTM-based model to predict the debater that makes the most convincing arguments. We find that incorporating argument structure features play an essential role in achieving the better predictive performance in assessing the persuasiveness of the arguments in online debates. 
\end{abstract}

\section{Introduction}

The increase in availability of online argumentation platforms has provided opportunity for researchers to develop computational methods at a larger scale studying the important factors of persuasiveness such as the language use \cite{hidey2017analyzing,tan2016winning,zhang2016conversational}, characteristics of audience (i.e. prior beliefs, demographics) \cite{durmus2019corpus,durmus2019exploring} and social interactions \cite{10.1145/3308558.3313676}. 

Prior work has showed incorporating argument structure features is important in assessing the quality of monological persuasive essays \cite{klebanov2016argumentation,wachsmuth2016using}. \citet{hidey2017analyzing} and \citet{egawa2019annotating} further collected annotations for semantic types of argument components and studied the relationship between the semantic types and persuasiveness of the arguments from online argumentative platform ChangeMyView (CMV) \cite{tan2016winning}. CMV consists of discussion trees where the users interact with the original poster to change their opinion on a given topic. Although the discussion trees are of a high quality since they are monitored by moderators \cite{tan2016winning}, they are not as structured since any user in the subreddit can participate in the discussions once the original post is posted. Furthermore, the persuasiveness of the posts in CMV is evaluated only by the original poster (i.e. whether they change their stance or not).  In this paper, we aim to investigate the effect of argument structure in persuasion on online debates. We focus on debates from DDO corpus \cite{durmus2019corpus} where debaters from two diverging sides of an issue express their opinions on a controversial topic in turns since these debates are more structured and the persuasiveness of the arguments in debates are evaluated by a larger set of audience. Moreover, this setup allows us to account for the  audience characteristics when studying the effect of the argument structure on persuasion. 
\begin{figure*}
\centering
\includegraphics[width=1.8\columnwidth]{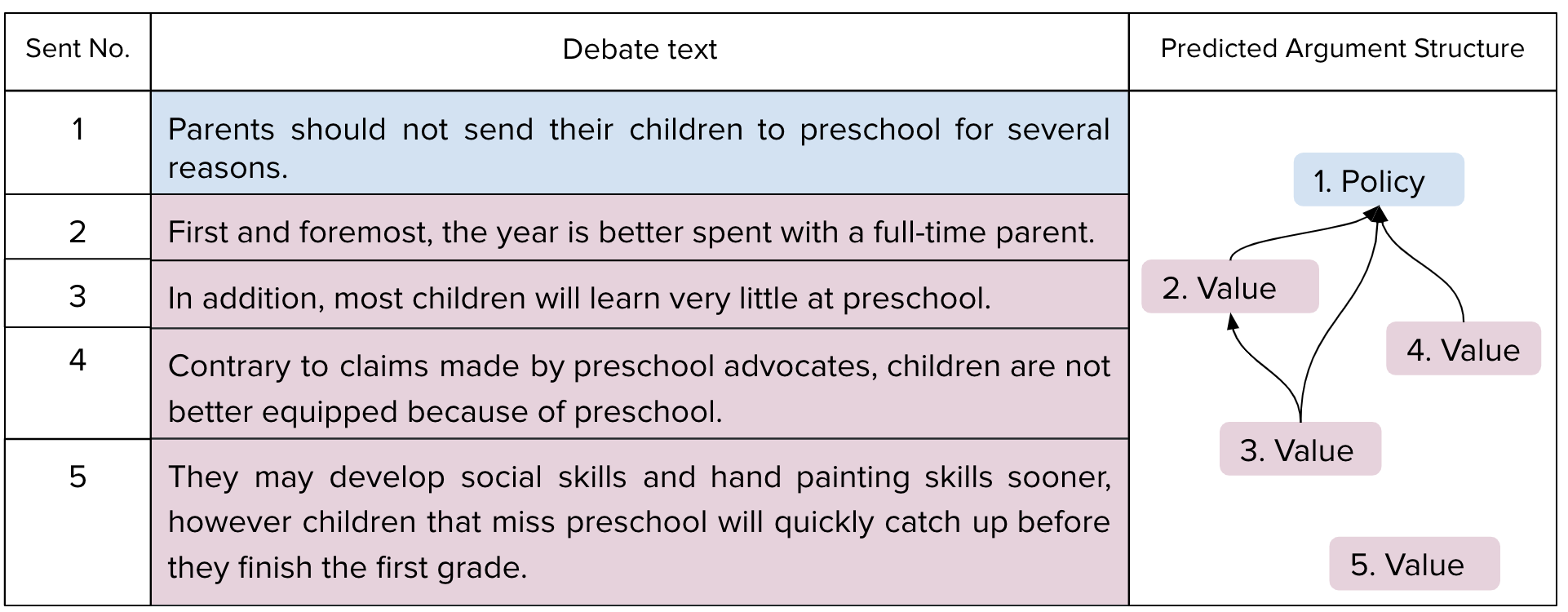}
\caption{An example of argument structure extracted from the debate text in one round from one side.}
\label{fig1}
\end{figure*}

We first generate argument structure on DDO dataset \cite{durmus2019corpus} using the model proposed by \newcite{niculae17marseille}. We then incorporate the features extracted from argument structure to an LSTM-based model that encodes the sequence of turns from two sides (i.e. {\sc pro} vs. {\sc con}). We compare our results with the baselines proposed in \cite{durmus2019exploring} which extracts linguistics features from the debate text as well as features that encode characteristics of the audience. We find that incorporating argument structure features achieves significantly better results than the baselines. Our analysis further shows that argument structure features encode important strategies of persuasion, for example, we find that more convincing arguments are more likely to include personal experiences of the debater and appeal to audience emotion.

\section{Related Work}\label{section2}

\textbf{Analysis of discourse structure}
There has been a lot effort to understand the role of discourse structure in argumentation. 
\citet{jiang2019applying} applied RST to essays written by students in K-12 schools and demonstrated its potential to provide automated feedback for essay quality. Argument structures, which can be considered as a special kind of discourse structure, have been widely analyzed in the task of automatic essay scoring and feedback \cite{klebanov2016argumentation,ghosh2016coarse,wachsmuth2016using}. Furthermore, \citet{duthie2018deep} has studied the relationship between ethos, a specific kind of argument unit, and the dynamics of governments from the UK parliamentary debates. The role of argument structure in persuasion on online debates is much less explored, which is the main focus of this paper.

\textbf{Analysis of Persuasion}
Prior studies on persuasion has mainly focused on understanding the role of linguistic factors \cite{petty1983central,chaiken1987heuristic,dillard2002persuasion,gold2015visual}. Besides, the interaction between debaters has shown to be an important cue in persuasion studies \cite{zhang2016conversational,tan2016winning,wang2017winning}. \citet{luu2019measuring} further found that the debater's skill estimated from debate text history is also predictive of convincing the audience. User factors are explored in previous papers \cite{durmus2019corpus,durmus2019exploring,longpre2019persuasion}, demonstrating the importance of characteristics and beliefs of the audience. Furthermore, \citet{potash2017towards} proposed a recurrent neural network architecture with attention and annotated audience favorability to predict the winner of the debate. \citet{villata2018assessing} and \citet{benlamine2017persuasive} studied the correlation of the engagement index in brain
hemispheres with the persuasion strategies. Argument structures have been used to understand argumentative strategies in dialogues and news editorials \cite{al2017patterns, wang2019persuasion}. A few studies have explored the impact of argument structures in predicting persuasion on CMV dataset based on statistical analysis of proposition types \cite{hidey2017analyzing,egawa2019annotating,morio2019revealing}. In this paper, we particularly study persuasion in online debates. We propose novel argument structure features based on n-grams of the supporting relations in argument structure graph of the debate text and experiment with these using both linear and neural models.

\section{Dataset}\label{section3.1}

We experiment with DDO dataset \cite{durmus2019corpus} which includes 77,655 debates covering 23 different topic categories. Each debate consists of multiple rounds with each round containing one utterance from the PRO side and one utterance from the CON side. Besides the text information for debates, the dataset also contains user information and votes provided by the audience on six different criteria of evaluating both the debaters. We use the criterion \textit{``Made more convincing arguments''} as an overall signal to study the role of argument structure in predicting more convincing arguments.

\begin{figure*}[t]
\centering
\includegraphics[width=2\columnwidth]{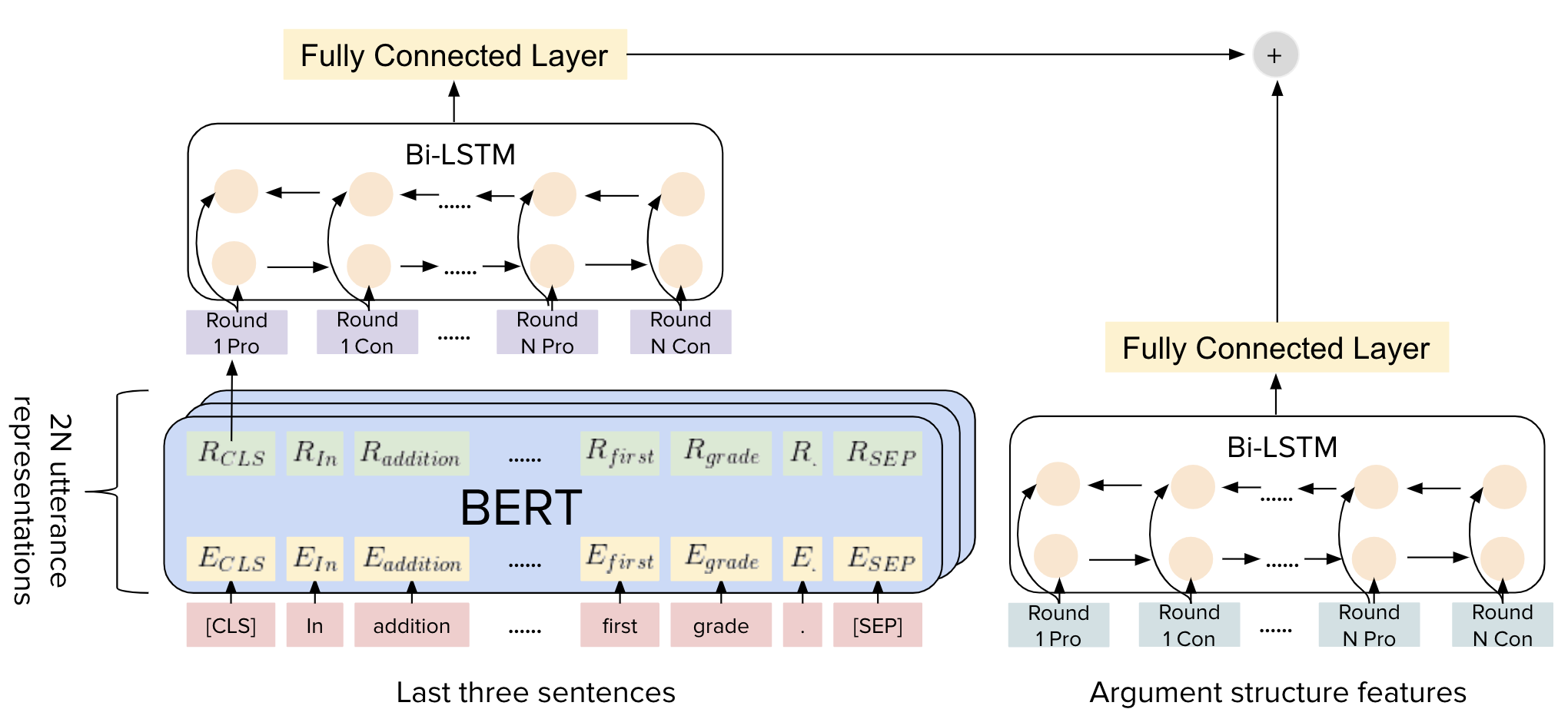}
\caption{Model for predicting which side makes more convincing arguments}
\label{fig2}
\end{figure*}

\section{Prediction Task}\label{section4}
\textbf{Task.} We aim to predict which side (i.e. {\sc pro} vs. {\sc con}) makes more convincing arguments during a debate, and thus is more persuasive. \\
\textbf{Data preprocessing.} We count which side of the debate gets more votes for the criterion \textit{``Made more convincing arguments''}. We eliminate debates if they are tied or the difference in votes is only 1.\footnote{Since the average number of total votes in one debate is 8, we consider difference of two or more votes as significant.}  The final dataset contains 2,606 debates.

\subsection{Argument structure features}

We apply the pre-trained model \cite{niculae17marseille} %
on DDO dataset to get the stucture of the arguments. We select this method since we can predict argumentative relations and classify proposition type at the same time, while the method proposed by \citet{chakrabarty2020ampersand} mainly focuses on predicting argumentative relations. Besides, this model can model argumentative relations that do not necessarily form a tree
structure which is more suitable to argumentation in the wild comparing to the models proposed in \newcite{stab2017parsing} and \newcite{peldszus2015joint}.
We generate argument structures for the selected 2,606 debates.\footnote{Since the model takes relatively long inference time and performs worse for long debates, we eliminate all the debates with more than 40 sentences in one round from one side. We also eliminate debates where one of the debaters forfeit during the debate.} The argument structure model outputs the proposition type for each sentence (i.e. {\sc reference}, {\sc testimony}, {\sc fact}, {\sc value}, {\sc policy}) as well as the supporting relationship between the propositions.
An example of argument structure generated on the text from one side in one round of the debate \textit{`Preschool Is A Waste Of Time'} is shown in Figure \ref{fig1}.  %
We use \textit{Amazon Mechanical Turk} (AMT) to further evaluate the quality of the argument structure on debates by asking Turkers to classify each argument from randomly picked 30 debates into five categories: {\sc Policy}, {\sc value}, {\sc fact}, {\sc testimony}, {\sc reference}. In total, we get annotations for 1,098 sentences, and each sentence is annotated by two annotators. %
We find that around 64\% of the output generated from the pre-trained model is consistent with either of the annotations from the Turkers.
We then extract three sets of argument structure features to capture the proposition types and link between propositions:

\textbf{Proposition n-gram frequency}
Similar to \newcite{wachsmuth2016using}, we obtain the frequencies of proposition unigrams, bigrams, and trigrams from the sequence of propositions. %
For example, ({\sc policy},{\sc value}) and  ({\sc value},{\sc value}) bigram features in Figure \ref{fig1} has values $0.25$,$0.75$ respectively. %

\textbf{Link n-gram frequency} We extract the n-gram information from the supporting relations in argument structure graph. For example, we represent two propositions connected with a link as a bigram (i.e $a \rightarrow b$ in the graph is represented with bigram (a,b)). 

\begin{table*}
\centering
\begin{tabular}{p{0.9\columnwidth}|p{0.9\columnwidth}}
\hline \multicolumn{1}{c}{\textbf{Pro}} & \multicolumn{1}{c}{\textbf{Con}} \\ \hline
[...] One of the quotes I remembered clearly was, ``God will give us whatever we want, as long as we don't screw up.'' [...] I haven't committed genocide or anything bad like that. But I've made my mistakes, and everyone has. [...] I'm not dead.  & [...] If cheating on test made someone happy, that doesn't make up for their unfair advantage. [...] Multiple times it is mentioned in the bible that homosexuality is wrong, it's a sin. ``You shall not lie with a male as one lies with a female; it is an abomination.'' [...]  \\ \hline
\end{tabular}
\caption{\label{table4} Example debate {\sc``Gay Marriage''} that is classified correctly after adding argument structure features. }
\end{table*}
\begin{table}[h]
\centering
\begin{tabular}{lc}
\hline \textbf{Model} & \textbf{Accuracy} \\ \hline
Majority baseline & 62.62\% \\ 
Linguistic+User LR & 67.41\% \\
Arg-Struct LR & 69.52\% \\
Linguistic+Arg-Struct LR & 70.48\% \\
Linguistic+User+Arg-Struct LR & 70.44\% \\
Our Model & \textbf{77.28\%} \\
Our Model w/o All Arg-Struct & 75.29\% \\
Our Model w/o Proposition N-gram & 76.21\% \\
Our Model w/o Link N-gram & 76.86\% \\ 
Our Model w/o Graphical & 76.95\% \\
\hline
\end{tabular}
\caption{\label{table1} Comparison with feature based Logistic Regression (LR). Arg-Struct denotes the argument structure features.}
\end{table}

\textbf{Graphical representation} \citet{rahwan2008mass} has found that there are five common argument structures in online environment: basic argument, convergent argument, serial argument, divergent argument, and linked argument. A typical basic argument is $a \rightarrow b$\footnote{$a$, $b$, $c$ denotes propositions and $a \rightarrow b$ denotes the directed link between $a$ and $b$.}, while serial argument is $a \rightarrow b\ \&\ b \rightarrow c$. The simplest convergent argument is in the form of $a \rightarrow b\ \&\ c \rightarrow b$, and a divergent argument is in the form of $a \rightarrow b\ \&\ a \rightarrow c$. Similarly, a linked argument is in the form of $a,c \rightarrow b$. %
We extract features to represent which of these types of arguments are used in the text of the debaters. We further classify the convergent arguments into two categories -- where two propositions support one proposition (regular convergent argument) and more than two propositions support one proposition (multi convergent argument). Similarly, we classify divergent argument into regular divergent argument and multi divergent argument. %

\subsection{Model Architecture}

We employ two separate bidirectional LSTM \cite{hochreiter1997long} models to encode the argument structure features and features encoding the debate text derived from pre-trained BERT model \cite{devlin-etal-2019-bert} as shown in Figure \ref{fig2}. %
LSTM modeling the debate text takes BERT representation \cite{devlin-etal-2019-bert} while LSTM encoding argument structure features takes three set of argument structure features of an utterance in a round at each time step.  
Two fully connected layers with softmax are used to predict the output probabilities over both of these LSTM models separately. The model learns weights during training to combine these probabilities.  %

\section{Experiments and Analysis}\label{section5}

We compare our model with the baseline proposed by \newcite{durmus2019corpus} employing linguistic features and features encoding audience characteristics.
The prediction accuracy is evaluated using 5-fold cross-validation, and the model parameters for each split are picked with 3-fold cross-validation on the training set. 
As shown in Table \ref{table1}, %
incorporating argument structure features to Logistic Regression achieves significantly better performance than the baseline with linguistic and user-based features. LSTM with argument structure features achieves the best predicive performance since LSTM can better represent context and the interplay between debaters. We perform t-test over 10 runs between the model with and without argument structure features, the $p$-value is 0.0038, indicating a statistically significant result. Furthermore, we do ablation over different sets of argument structure features. The results show that using the sequential flow of arguments is more effective than using argumentative relations in our setting. %

We further analyze what type of argument structure is more correlated with making more convincing arguments.
Comparing the unigram, bigram and trigram frequencies of the propositions by more convincing vs. less convincing debaters, we find that unigram {\sc testimony} ($p < 0.0001$)\footnote{The $p$-values are calculated using the Wilcoxon signed-rank test.}, bigram ({\sc value},{\sc testimony}) ($p < 0.001$), and trigram ({\sc value},{\sc testimony},{\sc value}) ($p < 0.0001$) appear more frequently in the more convincing side. This result suggests that justifying the objective claims with personal experiences is an effective strategy as also shown in previous work \cite{villata2018assessing}. Table \ref{table4} shows an example that is predicted classified by the model correctly after adding argument structure features. We observe that the side referring to their personal experiences ({\sc pro}) is voted as the side making more convincing arguments. 
Besides, we find that unigram {\sc Policy} ($p < 0.0001$), bigram ({\sc policy},{\sc value}) ($p < 0.005$) appear more frequently in the less convincing side suggesting that using propositions with type {\sc policy} -- which is used to specify a specific course of action to be taken --  may not be a very effective strategy in online debating.   %
Analyzing the link n-gram frequency features, we have further found that propositions with type {\sc value} from more convincing side are supported by a {\sc fact} ($p < 0.05$) more often. This suggests that the more convincing debaters may be using logos to support their views as also shown in previous work \cite{hidey2017analyzing}.  %
Finally, we observe that more convincing side tends to have more divergent arguments ($p=0.052$). %
Divergent arguments involves three or more consecutive sentences most of the time. In the case of three consecutive sentences, %
the middle sentence supports both the other two sentences by giving explanations or evidence, and serves as a transition between two similar ideas.

We also look into some examples that are classified wrong by the model. A typical error is caused by wrong proposition type classification. For example, in the debate ``Driving on public roads is a right not a privilege'', sentences from {\sc pro} side ``In addition, in purchasing our vehicles, we have the right to drive said vehicle.'' and ``I appreciate the insight given by my opponent but he/she has failed to address the issue at hand.'' are classified as ``Testimony'' wrongly, which makes the model prefer {\sc pro} as the more convincing side. We believe that incorporating more accurate argument structure generation models can further improve the performance on persuasion prediction.  

\section{Conclusion}\label{section6}
In this work, we explore the role of argument structure in online debate persuasion and find that incorporating argument structure features along with the linguistic features achieves the best predictive performance models. Moreover, we observe that argument structure features provide important cues about effective persuasion strategies in online debates. 

\section*{Acknowledgements}
We thank the anonymous reviewers for their useful feedback.
This work was supported in part by NSF grants IIS-1815455. The views and conclusions contained herein are those of the authors and should not be interpreted as necessarily representing the official policies or endorsements, either expressed or implied, of NSF or the U.S. Government.

\bibliographystyle{acl_natbib}
\bibliography{anthology,emnlp2020}

\begin{thebibliography}{39}
\expandafter\ifx\csname natexlab\endcsname\relax\def\natexlab#1{#1}\fi

\bibitem[{Al~Khatib et~al.(2017)Al~Khatib, Wachsmuth, Hagen, and
  Stein}]{al2017patterns}
Khalid Al~Khatib, Henning Wachsmuth, Matthias Hagen, and Benno Stein. 2017.
\newblock Patterns of argumentation strategies across topics.
\newblock In \emph{Proceedings of the 2017 Conference on Empirical Methods in
  Natural Language Processing}, pages 1351--1357.

\bibitem[{Benlamine et~al.(2017)Benlamine, Villata, Ghali, Frasson, Gandon, and
  Cabrio}]{benlamine2017persuasive}
Mohamed~S Benlamine, Serena Villata, Ramla Ghali, Claude Frasson, Fabien
  Gandon, and Elena Cabrio. 2017.
\newblock Persuasive argumentation and emotions: An empirical evaluation with
  users.
\newblock In \emph{International Conference on Human-Computer Interaction},
  pages 659--671. Springer.

\bibitem[{Chaiken(1987)}]{chaiken1987heuristic}
Shelly Chaiken. 1987.
\newblock The heuristic model of persuasion.
\newblock In \emph{Social influence: the ontario symposium}, volume~5, pages
  3--39. Hillsdale, NJ: Lawrence Erlbaum.

\bibitem[{Chakrabarty et~al.(2020)Chakrabarty, Hidey, Muresan, Mckeown, and
  Hwang}]{chakrabarty2020ampersand}
Tuhin Chakrabarty, Christopher Hidey, Smaranda Muresan, Kathy Mckeown, and
  Alyssa Hwang. 2020.
\newblock Ampersand: Argument mining for persuasive online discussions.
\newblock \emph{arXiv preprint arXiv:2004.14677}.

\bibitem[{Danescu-Niculescu-Mizil et~al.(2013)Danescu-Niculescu-Mizil, Sudhof,
  Jurafsky, Leskovec, and Potts}]{danescu2013computational}
Cristian Danescu-Niculescu-Mizil, Moritz Sudhof, Dan Jurafsky, Jure Leskovec,
  and Christopher Potts. 2013.
\newblock \href {https://www.aclweb.org/anthology/P13-1025} {A computational
  approach to politeness with application to social factors}.
\newblock In \emph{Proceedings of the 51st Annual Meeting of the Association
  for Computational Linguistics (Volume 1: Long Papers)}, pages 250--259,
  Sofia, Bulgaria. Association for Computational Linguistics.

\bibitem[{Devlin et~al.(2019)Devlin, Chang, Lee, and
  Toutanova}]{devlin-etal-2019-bert}
Jacob Devlin, Ming-Wei Chang, Kenton Lee, and Kristina Toutanova. 2019.
\newblock \href {https://doi.org/10.18653/v1/N19-1423} {{BERT}: Pre-training of
  deep bidirectional transformers for language understanding}.
\newblock In \emph{Proceedings of the 2019 Conference of the North {A}merican
  Chapter of the Association for Computational Linguistics: Human Language
  Technologies, Volume 1 (Long and Short Papers)}, pages 4171--4186,
  Minneapolis, Minnesota. Association for Computational Linguistics.

\bibitem[{Dillard and Pfau(2002)}]{dillard2002persuasion}
James~Price Dillard and Michael Pfau. 2002.
\newblock \emph{The persuasion handbook: Developments in theory and practice}.
\newblock Sage Publications.

\bibitem[{Duchi et~al.(2011)Duchi, Hazan, and Singer}]{duchi2011adaptive}
John Duchi, Elad Hazan, and Yoram Singer. 2011.
\newblock Adaptive subgradient methods for online learning and stochastic
  optimization.
\newblock \emph{Journal of Machine Learning Research}, 12(Jul):2121--2159.

\bibitem[{Durmus and Cardie(2018)}]{durmus2019exploring}
Esin Durmus and Claire Cardie. 2018.
\newblock \href {https://doi.org/10.18653/v1/N18-1094} {Exploring the role of
  prior beliefs for argument persuasion}.
\newblock In \emph{Proceedings of the 2018 Conference of the North {A}merican
  Chapter of the Association for Computational Linguistics: Human Language
  Technologies, Volume 1 (Long Papers)}, pages 1035--1045, New Orleans,
  Louisiana. Association for Computational Linguistics.

\bibitem[{Durmus and Cardie(2019{\natexlab{a}})}]{durmus2019corpus}
Esin Durmus and Claire Cardie. 2019{\natexlab{a}}.
\newblock \href {https://doi.org/10.18653/v1/P19-1057} {A corpus for modeling
  user and language effects in argumentation on online debating}.
\newblock In \emph{Proceedings of the 57th Annual Meeting of the Association
  for Computational Linguistics}, pages 602--607, Florence, Italy. Association
  for Computational Linguistics.

\bibitem[{Durmus and Cardie(2019{\natexlab{b}})}]{10.1145/3308558.3313676}
Esin Durmus and Claire Cardie. 2019{\natexlab{b}}.
\newblock \href {https://doi.org/10.1145/3308558.3313676} {Modeling the factors
  of user success in online debate}.
\newblock In \emph{The World Wide Web Conference}, WWW ’19, page 2701–2707,
  New York, NY, USA. Association for Computing Machinery.

\bibitem[{Duthie and Budzynska(2018)}]{duthie2018deep}
Rory Duthie and Katarzyna Budzynska. 2018.
\newblock A deep modular rnn approach for ethos mining.
\newblock In \emph{IJCAI}, pages 4041--4047.

\bibitem[{Egawa et~al.(2019)Egawa, Morio, and Fujita}]{egawa2019annotating}
Ryo Egawa, Gaku Morio, and Katsuhide Fujita. 2019.
\newblock Annotating and analyzing semantic role of elementary units and
  relations in online persuasive arguments.
\newblock In \emph{Proceedings of the 57th Annual Meeting of the Association
  for Computational Linguistics: Student Research Workshop}, pages 422--428.

\bibitem[{Feng and Hirst(2011)}]{feng2011classifying}
Vanessa~Wei Feng and Graeme Hirst. 2011.
\newblock Classifying arguments by scheme.
\newblock In \emph{Proceedings of the 49th annual meeting of the association
  for computational linguistics: Human language technologies}, pages 987--996.

\bibitem[{Ghosh et~al.(2016)Ghosh, Khanam, Han, and Muresan}]{ghosh2016coarse}
Debanjan Ghosh, Aquila Khanam, Yubo Han, and Smaranda Muresan. 2016.
\newblock Coarse-grained argumentation features for scoring persuasive essays.
\newblock In \emph{Proceedings of the 54th Annual Meeting of the Association
  for Computational Linguistics (Volume 2: Short Papers)}, pages 549--554.

\bibitem[{Gold et~al.(2015)Gold, El-Assady, Hautli-Janisz, B{\"o}gel,
  Rohrdantz, Butt, Holzinger, and Keim}]{gold2015visual}
Valentin Gold, Mennatallah El-Assady, Annette Hautli-Janisz, Tina B{\"o}gel,
  Christian Rohrdantz, Miriam Butt, Katharina Holzinger, and Daniel Keim. 2015.
\newblock Visual linguistic analysis of political discussions: Measuring
  deliberative quality.
\newblock \emph{Digital Scholarship in the Humanities}, 32(1):141--158.

\bibitem[{Hidey et~al.(2017)Hidey, Musi, Hwang, Muresan, and
  McKeown}]{hidey2017analyzing}
Christopher Hidey, Elena Musi, Alyssa Hwang, Smaranda Muresan, and Kathleen
  McKeown. 2017.
\newblock Analyzing the semantic types of claims and premises in an online
  persuasive forum.
\newblock In \emph{Proceedings of the 4th Workshop on Argument Mining}, pages
  11--21.

\bibitem[{Hochreiter and Schmidhuber(1997)}]{hochreiter1997long}
Sepp Hochreiter and J{\"u}rgen Schmidhuber. 1997.
\newblock Long short-term memory.
\newblock \emph{Neural computation}, 9(8):1735--1780.

\bibitem[{Jiang et~al.(2019)Jiang, Yang, Suvarna, Casula, Zhang, and
  Rose}]{jiang2019applying}
Shiyan Jiang, Kexin Yang, Chandrakumari Suvarna, Pooja Casula, Mingtong Zhang,
  and Carolyn Rose. 2019.
\newblock Applying rhetorical structure theory to student essays for providing
  automated writing feedback.
\newblock In \emph{Proceedings of the Workshop on Discourse Relation Parsing
  and Treebanking 2019}, pages 163--168.

\bibitem[{Klebanov et~al.(2016)Klebanov, Stab, Burstein, Song, Gyawali, and
  Gurevych}]{klebanov2016argumentation}
Beata~Beigman Klebanov, Christian Stab, Jill Burstein, Yi~Song, Binod Gyawali,
  and Iryna Gurevych. 2016.
\newblock Argumentation: Content, structure, and relationship with essay
  quality.
\newblock In \emph{Proceedings of the Third Workshop on Argument Mining
  (ArgMining2016)}, pages 70--75.

\bibitem[{Krippendorff(1970)}]{krippendorff1970estimating}
Klaus Krippendorff. 1970.
\newblock Estimating the reliability, systematic error and random error of
  interval data.
\newblock \emph{Educational and Psychological Measurement}, 30(1):61--70.

\bibitem[{Longpre et~al.(2019)Longpre, Durmus, and
  Cardie}]{longpre2019persuasion}
Liane Longpre, Esin Durmus, and Claire Cardie. 2019.
\newblock Persuasion of the undecided: Language vs. the listener.
\newblock In \emph{Proceedings of the 6th Workshop on Argument Mining}, pages
  167--176.

\bibitem[{Luu et~al.(2019)Luu, Tan, and Smith}]{luu2019measuring}
Kelvin Luu, Chenhao Tan, and Noah~A Smith. 2019.
\newblock Measuring online debaters’ persuasive skill from text over time.
\newblock \emph{Transactions of the Association for Computational Linguistics},
  7:537--550.

\bibitem[{Morio et~al.(2019)Morio, Egawa, and Fujita}]{morio2019revealing}
Gaku Morio, Ryo Egawa, and Katsuhide Fujita. 2019.
\newblock Revealing and predicting online persuasion strategy with elementary
  units.
\newblock In \emph{Proceedings of the 2019 Conference on Empirical Methods in
  Natural Language Processing and the 9th International Joint Conference on
  Natural Language Processing (EMNLP-IJCNLP)}, pages 6275--6280.

\bibitem[{Niculae et~al.(2017)Niculae, Park, and Cardie}]{niculae17marseille}
Vlad Niculae, Joonsuk Park, and Claire Cardie. 2017.
\newblock {Argument Mining with Structured SVMs and RNNs}.
\newblock In \emph{Proceedings of ACL}.

\bibitem[{Peldszus and Stede(2015)}]{peldszus2015joint}
Andreas Peldszus and Manfred Stede. 2015.
\newblock Joint prediction in mst-style discourse parsing for argumentation
  mining.
\newblock In \emph{Proceedings of the 2015 Conference on Empirical Methods in
  Natural Language Processing}, pages 938--948.

\bibitem[{Petty et~al.(1983)Petty, Cacioppo, and Schumann}]{petty1983central}
Richard~E Petty, John~T Cacioppo, and David Schumann. 1983.
\newblock Central and peripheral routes to advertising effectiveness: The
  moderating role of involvement.
\newblock \emph{Journal of consumer research}, 10(2):135--146.

\bibitem[{Potash and Rumshisky(2017)}]{potash2017towards}
Peter Potash and Anna Rumshisky. 2017.
\newblock Towards debate automation: a recurrent model for predicting debate
  winners.
\newblock In \emph{Proceedings of the 2017 Conference on Empirical Methods in
  Natural Language Processing}, pages 2465--2475.

\bibitem[{Rahwan(2008)}]{rahwan2008mass}
Iyad Rahwan. 2008.
\newblock Mass argumentation and the semantic web.
\newblock \emph{Web Semantics: Science, Services and Agents on the World Wide
  Web}, 6(1):29--37.

\bibitem[{Somasundaran et~al.(2007)Somasundaran, Ruppenhofer, and
  Wiebe}]{somasundaran2007detecting}
Swapna Somasundaran, Josef Ruppenhofer, and Janyce Wiebe. 2007.
\newblock Detecting arguing and sentiment in meetings.
\newblock In \emph{Proceedings of the SIGdial Workshop on Discourse and
  Dialogue}, volume~6.

\bibitem[{Stab and Gurevych(2017)}]{stab2017parsing}
Christian Stab and Iryna Gurevych. 2017.
\newblock Parsing argumentation structures in persuasive essays.
\newblock \emph{Computational Linguistics}, 43(3):619--659.

\bibitem[{Tan and Lee(2016)}]{tan2016talk}
Chenhao Tan and Lillian Lee. 2016.
\newblock Talk it up or play it down?(un) expected correlations between (de-)
  emphasis and recurrence of discussion points in consequential us economic
  policy meetings.
\newblock \emph{arXiv preprint arXiv:1612.06391}.

\bibitem[{Tan et~al.(2016)Tan, Niculae, Danescu-Niculescu-Mizil, and
  Lee}]{tan2016winning}
Chenhao Tan, Vlad Niculae, Cristian Danescu-Niculescu-Mizil, and Lillian Lee.
  2016.
\newblock Winning arguments: Interaction dynamics and persuasion strategies in
  good-faith online discussions.
\newblock In \emph{Proceedings of the 25th international conference on world
  wide web}, pages 613--624. International World Wide Web Conferences Steering
  Committee.

\bibitem[{Villata et~al.(2018)Villata, Benlamine, Cabrio, Frasson, and
  Gandon}]{villata2018assessing}
Serena Villata, Sahbi Benlamine, Elena Cabrio, Claude Frasson, and Fabien
  Gandon. 2018.
\newblock Assessing persuasion in argumentation through emotions and mental
  states.
\newblock In \emph{The Thirty-First International Flairs Conference}.

\bibitem[{Wachsmuth et~al.(2016)Wachsmuth, Al~Khatib, and
  Stein}]{wachsmuth2016using}
Henning Wachsmuth, Khalid Al~Khatib, and Benno Stein. 2016.
\newblock Using argument mining to assess the argumentation quality of essays.
\newblock In \emph{Proceedings of COLING 2016, the 26th International
  Conference on Computational Linguistics: Technical Papers}, pages 1680--1691.

\bibitem[{Wang et~al.(2017)Wang, Beauchamp, Shugars, and Qin}]{wang2017winning}
Lu~Wang, Nick Beauchamp, Sarah Shugars, and Kechen Qin. 2017.
\newblock Winning on the merits: The joint effects of content and style on
  debate outcomes.
\newblock \emph{Transactions of the Association for Computational Linguistics},
  5:219--232.

\bibitem[{Wang et~al.(2019)Wang, Shi, Kim, Oh, Yang, Zhang, and
  Yu}]{wang2019persuasion}
Xuewei Wang, Weiyan Shi, Richard Kim, Yoojung Oh, Sijia Yang, Jingwen Zhang,
  and Zhou Yu. 2019.
\newblock \href {https://doi.org/10.18653/v1/P19-1566} {Persuasion for good:
  Towards a personalized persuasive dialogue system for social good}.
\newblock In \emph{Proceedings of the 57th Annual Meeting of the Association
  for Computational Linguistics}, pages 5635--5649, Florence, Italy.
  Association for Computational Linguistics.

\bibitem[{Wilson et~al.(2005)Wilson, Wiebe, and
  Hoffmann}]{wilson2005recognizing}
Theresa Wilson, Janyce Wiebe, and Paul Hoffmann. 2005.
\newblock Recognizing contextual polarity in phrase-level sentiment analysis.
\newblock In \emph{Proceedings of Human Language Technology Conference and
  Conference on Empirical Methods in Natural Language Processing}.

\bibitem[{Zhang et~al.(2016)Zhang, Kumar, Ravi, and
  Danescu-Niculescu-Mizil}]{zhang2016conversational}
Justine Zhang, Ravi Kumar, Sujith Ravi, and Cristian Danescu-Niculescu-Mizil.
  2016.
\newblock \href {https://doi.org/10.18653/v1/N16-1017} {Conversational flow in
  {O}xford-style debates}.
\newblock In \emph{Proceedings of the 2016 Conference of the North {A}merican
  Chapter of the Association for Computational Linguistics: Human Language
  Technologies}, pages 136--141, San Diego, California. Association for
  Computational Linguistics.

\end{thebibliography}

\newpage
\appendix
\section{Appendix}

\subsection{Argument Structure Features Used}
\textbf{Proposition n-gram frequency} When we eliminate proposition bigram and trigram that occur less than 3\% in all training debates, five types of unigrams, eight types of bigrams and ten types of trigrams remain.  \\ 
\textbf{Unigram types:} policy, value, fact, testimony, reference.\\
\textbf{Bigram types:} (value, value), (testimony, value), (value, testimony), (value, policy), (policy, value), (fact, value), (value, fact), (testimony, testimony). \\ 
\textbf{Trigram types}: (value, value, value), (testimony, value, value), (value, value, policy), (value, value, testimony), (value, testimony, value), (fact, value, value), (policy, value, value), (value, fact, value), (value, policy, value), (value, value, fact).\\
\textbf{Link n-gram frequency} When we eliminate all link bigrams that occur less than 3\% of all training data, four types of link bigrams remain (i.e. (value, value), (value, policy), (fact, value), (testimony, value)).\\
\textbf{Graphical representation} There are 5 types of features for graphical representation: basic argument, regular convergent argument, regular divergent argument, multi convergent argument, multi divergent argument. 

In total, the argument structure features are 32-dimensional.

\subsection{Linguistic Features and User Features}
The linguistic features and user features we use for the Logistic Regression based baseline is the same as the features used by \newcite{durmus2019corpus}. They include hedge words \cite{tan2016talk}, evidence words (e.g. ``according to''), positive words, negative words, swear words, personal pronouns, tf-idf, argument lexicon
features \cite{somasundaran2007detecting}, politeness marks \cite{danescu2013computational},
sentiment, connotation \cite{feng2011classifying},
subjectivity \cite{wilson2005recognizing}, modal verbs, type-token ratio (diverse word usage), and punctuation.

The user features include opinion similarity for big issues, religious and political ideology match and persuadability score (how likely a person will be persuaded) \cite{longpre2019persuasion}.

\begin{figure*}[h]
\centering
\includegraphics[width=1.8\columnwidth]{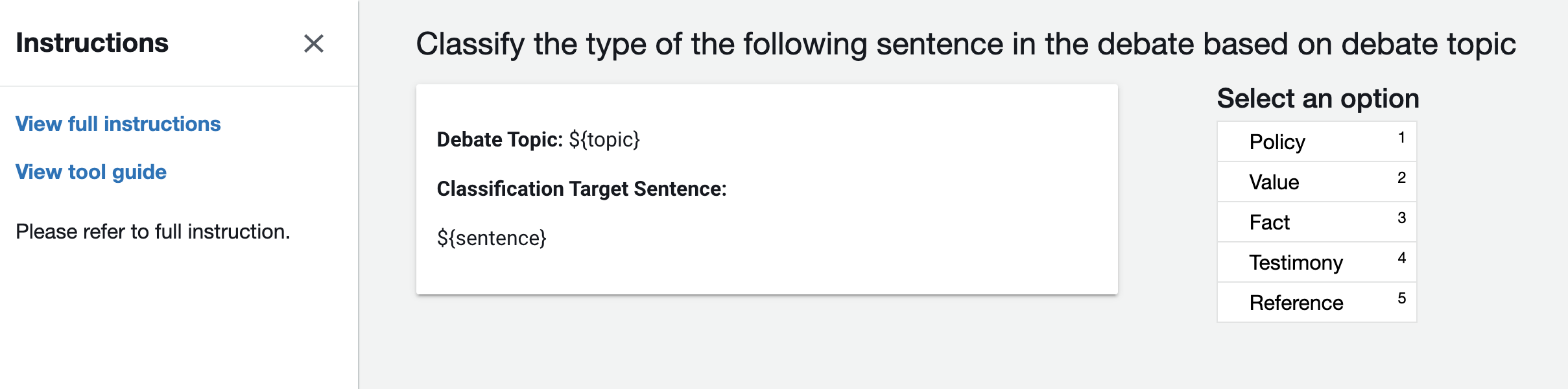}
\caption{AMT annotation example}
\label{fig4}
\end{figure*}
\subsection{BERT Representation Generation}
We input the utterance in one round for one debater. The segment embedding for each word in the utterance is the same, though one utterance will contain multiple sentences. Due to the maximum sequence length of 128 tokens of BERT \footnote{We use BERT-base with uncased input as the pretrained model.}, which is much shorter than the average length of utterance in one round for one debater, we truncate the debate text input and only preserve the last three sentences\footnote{We also experimented using more sentences (e.g. last five sentences) in cases where the sequence length has not been maxed out has also been tested, but it doesn't show significant improvement.} in each round for each debater. The truncate method of choosing the first three sentences of the utterance has also been tested, but the performance of the model was around 3\% lower.

\subsection{Implementation Details}
We use grid search to pick the hyperparameter. For the model that encodes linguistic information, we use a one-layer bidirectional LSTM with 768 dimension BERT representation input and 32 dimension hidden states. (We search in [16, 32, 64] for hidden dimension.) For the model that encodes argument structure information, we use a one-layer bidirectional LSTM with 32 dimension argument features input and 4 dimension hidden states. (We search in [16, 8, 4] for hidden dimension). We have a 0.5 dropout rate for both fully connected layers. Total number of parameters is around 100k. We use Adagrad \cite{duchi2011adaptive} with initial learning rate 0.005 and weight decay 0.01 to optimize the cross-entropy loss. (We also experiment with Adam with default setting, Adagrad without weight decay, learning rate between [0.001, 0.005, 0.01]). 2200 debates are used for training, 200 for validation and 206 for test set. We use early stopping to avoid overfitting, the model is trained for around 15 epochs on average. It takes less than 15 minutes to run the model on a CPU (2.7 GHz Intel Core i7). To test the stability of our results, we train and evaluate our model 10 times and take the average accuracy.
\begin{table}[]
\centering
\begin{tabular}{lcc}
\hline \textbf{Type} & \textbf{\# Proposition} &\textbf{Consistency} \\ \hline
Policy & 97 &56.70\% \\ 
Value & 834 & 65.47\% \\
Fact & 85 & 84.71\% \\
Testimony & 79 & 37.97\% \\
Reference & 3 & 33.33\% \\
All & 1,098 & 64.12\%\\
\hline
\end{tabular}
\caption{\label{table5} Annotation results from Amazon Mechanical Turker.}
\end{table}

\subsection{Details on AMT result}

Figure \ref{fig4} shows the screenshot for a typical HIT for the Turkers. For each HIT, the turkers are given the debate topic and the sentence to be classified. They need to choose between 5 categories: Policy, Value, Fact, Testimony, Reference. The definition of these proposition types and the corresponding example are included in the full instruction.

In total, we get annotations for 1,098 sentences from seventeen annotators. The detailed results are listed in Table \ref{table5}. Consistency means the generated annotations is consistent with either of the annotations from the Turkers. We also compute Inter-Annotator Agreement (IAA) using Kripendorff’s alpha \cite{krippendorff1970estimating}. The Kripendorff's alpha is 0.2, indicating that annotating argument structure is still a hard task for Turkers.

\end{document}